\documentclass{egpubl}
\pdfoutput=1
\usepackage{pg2021}
 
\SpecialIssuePaper         

\usepackage[T1]{fontenc}
\usepackage{dfadobe}  

\usepackage{cite}  
\BibtexOrBiblatex
\electronicVersion
\PrintedOrElectronic
\ifpdf \usepackage[pdftex]{graphicx} \pdfcompresslevel=9
\else \usepackage[dvips]{graphicx} \fi

\usepackage{egweblnk} 

\usepackage[caption=false]{subfig}
\usepackage{array}
\newcolumntype{M}[1]{>{\centering\arraybackslash}m{#1}}
\usepackage{amsmath}
\usepackage{pdfpages}

\usepackage{xcolor}

\newcommand{\rev}[1]{{\color{black}{#1}}}

\newcommand{\degree}{$^{\circ}$}
\renewcommand{\vec}[1]{\mathbf{#1}}

\title[Interactive Analysis of CNN Robustness]%
      {Interactive Analysis of CNN Robustness}

\author[Sietzen, S., Lechner, M., Borowski, J., Hasani, R., Waldner, M.]
{\parbox{\textwidth}{\centering Stefan Sietzen$^{1}$, 
        Mathias Lechner$^{2}$\orcid{0000-0002-6117-0076}, 
        Judy Borowski$^{3}$\orcid{0000-0002-3746-7152}, 
        Ramin Hasani$^{1,4}$\orcid{0000-0002-9889-5222}, 
        and Manuela Waldner$^{1}$\orcid{0000-0003-1387-5132} 
        }
        \\
{\parbox{\textwidth}{\centering $^1$TU Wien, Vienna, Austria\\
         $^2$ Institute of Science and Technology Austria (IST Austria), Klosterneuburg, Austria\\
         $^3$ University of Tübingen, Germany \\
        $^4$ Massachusetts Institute of Technology (MIT), Cambridge, USA \\
      } 
}
}

\begin{document}
\null
\includepdf[fitpaper=true, pages=-]{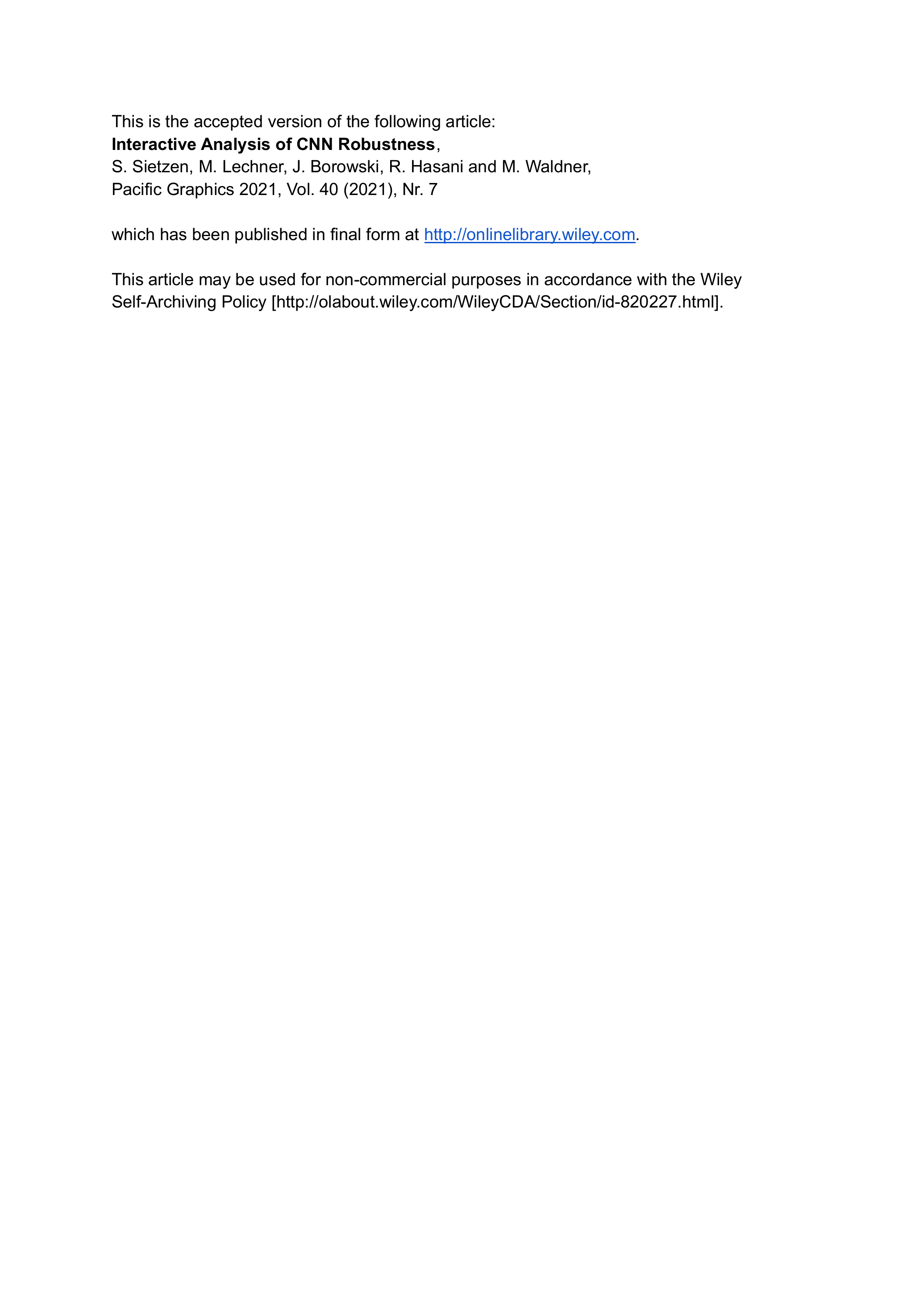}

\teaser{
 \includegraphics[width=0.9\linewidth]{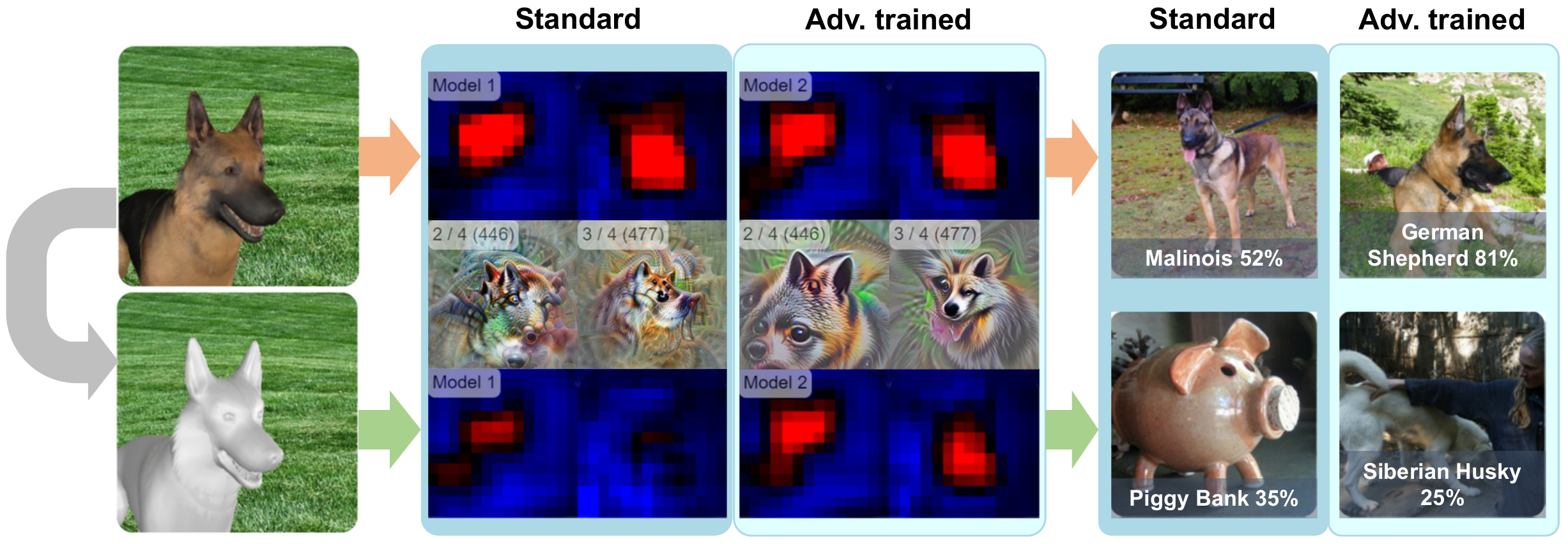}
 \centering
  \caption{Rapid interactive comparison of two models' network responses through activations of two selected ``dog-relevant'' neurons (middle) and predicted target classes (right) to input scene perturbations of a dog (left): Removing the texture leaves the 3D model with a smooth surface and causes a standard model to identify a piggy bank instead of a dog, while an adversarially trained model identifies a white dog breed (Siberian Husky). Activations of dog-relevant neurons in the standard model decrease due to missing texture information. }
\label{fig:teaser}
}

\maketitle
\begin{abstract}
While convolutional neural networks (CNNs) have found wide adoption as state-of-the-art models for image-related tasks, their predictions are often highly sensitive to small input perturbations, which the human vision is robust against.
This paper presents Perturber, a web-based application that allows users to instantaneously explore how CNN activations and predictions evolve when a 3D input scene is interactively perturbed. Perturber offers a large variety of scene modifications, such as camera controls, lighting and shading effects, background modifications, object morphing, as well as adversarial attacks, to facilitate the discovery of potential vulnerabilities. Fine-tuned model versions can be directly compared for qualitative evaluation of their robustness.  
Case studies with machine learning experts have shown that Perturber helps users to quickly generate hypotheses about model vulnerabilities and to qualitatively compare model behavior.
Using quantitative analyses, we could generalize users' insights to other CNN architectures and input images, yielding new insights about the vulnerability of adversarially trained models. 

\ccsdesc[500]{Human-centered computing~Visualization systems and tools}
\ccsdesc[500]{Computing methodologies~Machine learning}

\printccsdesc   
\end{abstract}  
\section{Introduction}

Convolutional neural networks (CNNs) achieve impressive results in a variety of applications, such as image classification~\cite{krizhevsky_imagenet_2017} or object detection~\cite{ren_faster_2017}. The performance of a CNN trained on a given training data set is typically assessed in terms of prediction accuracy on a held-out validation dataset. If the statistical distributions of the training and validation set differ, the high performance can drop precipitously. In that case, the model is not robust against the variability within the validation data. CNN robustness has been shown to be affected by image perturbations such as cropping~\cite{zheng2016improving}, blur~\cite{dodge2016understanding}, high-frequency noise~\cite{yin_fourier_2019}, or texture variations~\cite{geirhos_imagenet_2019}. Prominent examples of noise perturbation are adversarial attacks: Single pixels of an input image are changed slightly such that a CNN misclassifies it~\cite{szegedy_intriguing_2014}. To humans, these changes are typically imperceptible, and they would still assign the correct labels. 

Robustness is a highly safety-critical aspect of CNNs in various applications, such as self-driving cars~\cite{lechner2020neural}. For example, researchers have demonstrated how targeted yet unsuspicious changes of traffic signs can cause CNNs to consistently miss stop signs~\cite{eykholt_robust_2018}. Understanding which factors influence the robustness of CNNs and, consequently, designing and evaluating more robust models are therefore research topics of central importance in the machine learning community~\cite{goodfellow2014explaining,szegedy_intriguing_2014,madry_towards_2019,lechner2021adversarial}. 

To create a basis for tackling robustness, researchers aim to gain a better general understanding of which image features CNNs are most sensitive to and how this sensitivity differs from human visual perception \cite{babaiee2021off}. For example, Geirhos et al.~\cite{geirhos_imagenet_2019} investigated the influence of shape versus texture on humans and CNNs in image classification and found a strong bias for texture in CNNs. This texture bias was also confirmed by Zhang and Zhu~\cite{zhang_interpreting_2019} by evaluating the effect of image saturation or random shuffling of image patches. Through systematic analysis of image perturbation in the Fourier domain, Yin et al.~\cite{yin_fourier_2019} could show that CNNs are highly sensitive to high-frequency perturbations. 

A common way to improve robustness is to train models on more variable input images, using data augmentation methods~\cite{perez2017effectiveness}, image stylization~\cite{geirhos_imagenet_2019}, or adversarial examples~\cite{madry_towards_2019}. 
To evaluate the performance of these improved models, they are then compared to standard CNN models. 
To perform these analyses, researchers automatically perturb images offline, log the predictions for each perturbed input image, and compare the results between models. 
According to our collaborating domain experts, such experiments take several hours to set up and include time-consuming parameter searches. For some experiments, it is not even possible to determine network performance fully automatically as the ground truth of the input images also becomes unclear for humans because of the performed image perturbations. Clearly, these factors severely limit machine learning experts to quickly explore factors of image perturbation that may impact CNN performance. 


We hypothesize that \emph{being able to interactively manipulate a synthetic input scene with a large and diverse set of visual perturbation parameters and observing the changing activations and predictions instantly will allow researchers to quickly generate hypotheses and build a stronger intuition for potential vulnerabilities of CNNs}. 
In this work, we therefore introduce Perturber -- a novel real-time experimentation interface for exploratory analysis of model robustness (see Figure~\ref{fig:teaser}). 
Our work provides the following contributions: 

\begin{enumerate}
    \item Interactive input perturbations of 3D scenes in combination with feature visualizations, activation maps, and model predictions as a novel approach to interactively explore model robustness. 
    \item A direct comparison interface for qualitative visual validation of more robust, fine-tuned model variants.  
    \item Implementation of a publicly accessible web-based VA tool\footnote{\url{http://perturber.stefansietzen.at/}} that supports a large variety of perturbation methods of 3D input scenes and visualizes the model responses in real-time. 
    \item Observations from case studies with machine learning experts demonstrating that live inspection of input perturbations allows experts to visually explore known vulnerabilities, to compare model behaviors, and to generate new hypotheses concerning model robustness. 
    \item Quantitative verification of selected user observations from the case study shedding new light on the robustness of adversarially trained models. 
\end{enumerate}

\section{Related Work}

In recent years, a wide spectrum of deep learning visualization methods has emerged. 
For a comprehensive overview of visual analytics (VA) for deep learning, we refer the reader to a survey by Hohman et al.~\cite{hohman_visual_2019}.
For example, graph structure visualizations, such as the \emph{TensorFlow Graph Visualizer}~\cite{wongsuphasawat_visualizing_2018}, help users to get a better understanding of their models' structure, which comprise numerous layers and connections. 
 Others, such as \emph{DeepEyes}~\cite{pezzotti_deepeyes_2018} and \emph{DeepTracker}~\cite{liu_deeptracker_2019}, track detailed metrics throughout the training process to facilitate the identification of model problems or anomalous iterations. 
 \emph{ExplAIner}~\cite{spinner_explainer_2020} goes beyond monitoring and also integrates different steering mechanisms to help users understand and optimize their models. 
 A case study using a VA system to assess a model's performance to detect and classify traffic lights has shown that interactive VA systems can successfully guide experts to improve their training data~\cite{gou2020vatld}. 

While these examples all focus on the inspection of a single model, others support model comparisons. 
For example, using \emph{REMAP}~\cite{cashman_ablate_2020}, users can rapidly create model architectures through ablations (i.e., removing single layers of an existing model) and variations (i.e., creating new models through layer replacements) and compare the created models by their structure and performance.
Ma et al.~\cite{ma_visual_2020} designed multiple coordinated views to help experts analyze network model behaviors after transfer learning. 
\emph{CNN\-Comparator}~\cite{zeng2017cnncomparator} uses multiple coordinated views to compare the architecture and the prediction of a selected input image between two CNNs. Model comparison is also a crucial aspect of our work. However, the focus of the present work lies on fluid modification of input stimuli and instantaneous analysis and comparison of the network responses. Thus, we let users \emph{generate and perturb} input images from 3D scenes in a ``playground-like'' manner rather than letting the user select input images with a ground-truth class label. 

Interactive ``playgrounds'' require relatively little underlying deep learning knowledge and can be used for educational purposes. For more informed users, they are valuable for building an intuition and validating knowledge from literature~\cite{kahng_gan_2019}. 
Notable examples are \emph{TensorFlow Playground} \cite{smilkov_direct-manipulation_2017}, which supports the interactive modification and training of DNNs in the browser, and \emph{GANLab}~\cite{kahng_gan_2019}, which is designed in a similar style and supports experimentation with Generative Adversarial Networks. \emph{CNN Explainer}~\cite{wang_cnn_2020} provides a visual explanation of the inner workings of a CNN by showing connections between layers and activation maps, allowing the user to choose the input from a fixed collection of images. 
More similarly to our work, Harley~\cite{bebis_interactive_2015} provides an online tool where users can draw digits onto a canvas. Then, the responses of all neurons in a simple MNIST-trained network are visualized in real-time. \emph{Adversarial Playground}~\cite{norton2017adversarial} allows users to interactively generate adversarial attacks and instantly observe the predictions of a simple MNIST-trained network. 
To support interactive probing of model responses based on input modifications, \emph{Prospector}~\cite{krause_interacting_2016}, the \emph{what-if-tool}~\cite{wexler_what-if_2020}, and \emph{NLIZE}~\cite{liu2018nlize} allow users to interrogate the model by varying the input in the domains of tabular data and natural language processing, respectively. 
These works inspired us to build a system that lets users \emph{interactively explore} model robustness through input perturbations. In contrast to prior work, Perturber operates on complex CNN models, such as Inception-V1 trained on ImageNet, and can be used to discover and explain vulnerabilities to complex input scenes, like animals or man-made objects in different environments. 

To explain a CNN model, there are powerful methods to reveal the role of a network's hidden units by visualizing their learned features. \emph{Feature visualization} is an activation maximization 
technique, which was improved by combining a variety of regularization techniques~\cite{yosinski_understanding_2015, olah_feature_2017}. Feature visualizations have been used to 
identify and characterize causal connections of neurons in CNNs~\cite{cammarata_thread_2020}, and to comprehensively document the role of individual neurons in large CNNs \cite{OpenAIMi10:online}. Within VA tools, feature visualizations have been used to compare learned features before and after transfer learning~\cite{ma_visual_2020} or to visualize a graph of the most relevant neurons and their connections for a selected target class~\cite{hohman_summit_2020}. Similarly, \emph{Bluff}~\cite{das_bluff_2020} shows a graph containing the most relevant neurons explaining precomputed adversarial attacks, where neurons are represented by their feature visualizations.

Other powerful interpretability methods are \emph{saliency maps} (or \emph{attribution maps}), which show the saliency of the input image's regions with respect to the selected target class or network component~\cite{simonyan2013deep}. 
Saliency maps and other gradient-based methods like \emph{LRP} \cite{lapuschkin_lrp_2016}, \emph{Integrated Gradients} \cite{sundararajan_axiomatic_2017}, or \emph{Grad-CAM} \cite{selvaraju_grad-cam_2017}, however, require a computationally expensive back-propagation pass. A very simple solution to reveal relevant image regions for a model's prediction is to directly visualize the forward-propagation activations of selected feature maps in intermediate layers. For example, the \emph{DeepVis Toolbox}~\cite{yosinski_understanding_2015} shows live visualizations of CNN activations from a webcam feed. 
The goal is to get a general intuition what features a CNN has learned. 
\emph{AEVis}~\cite{liu_analyzing_noise_2018} shows activations of neurons to a pre-defined set of input images along a ``datapath visualization''. This visualization allows users to trace the effects of adversarial attacks through the hidden layers of a network. Datapaths are formed by critical neurons and their connections that are responsible for the predictions. 
Like the \emph{DeepVis Toolbox}~\cite{yosinski_understanding_2015}, Perturber visualizes activations and predictions based on live input. 
The major difference is that Perturber generates input images from an interactive 3D scene and provides a rich palette of input perturbation methods to gradually explore potential vulnerabilities of a model. In addition, it facilitates direct comparison between a standard model and a more robust variation thereof to explore the benefits and limitations of model variations. 
\section{Perturber Interface}

The high-level goal of Perturber is to interactively explore potential sources of vulnerabilities for CNNs to facilitate the design of more robust models. 
Through constant exchange between visualization researchers and machine learning experts investigating model interpretability and robustness,  we identified four central requirements of a VA application to support exploratory analysis and qualitative validation of model robustness: 

\textbf{R1:} Rich \textbf{online input perturbation} of a representative scene is essential to quickly generate input images for which model responses can be investigated. The system should provide a large variety of possible perturbation parameters and allow a flexible combination of these perturbations. 

\textbf{R2:} To understand what makes a model robust, it is not sufficient to understand \emph{how} a model responds to the perturbed input but also \emph{why} the model's responses change. To this end, looking not only at the input and output, but particularly at the numerous \textbf{hidden layers} is inevitable 
when trying to understand what makes a model react unpredictably for humans.

\textbf{R3:} Interactivity can help users to build intuitions through dynamic experiments~\cite{kahng2019does}. We thus aim to make the model responses to input perturbations \textbf{instantly} visible to support a fluid feedback loop in a playground-like environment. 
     
\textbf{R4:} 
Robustness can be achieved by training models on more versatile input images, using data augmentation methods, such as affine image transformations, image stylization, or adversarial training. 
A direct visual \textbf{comparison} between the standard CNN model and its more robust fine-tuned version are necessary for a first qualitative validation whether a fine-tuned model is generally more robust to input perturbations or only selectively more robust to specific perturbations it has been trained on.

\begin{figure*}[h]
\centering
\includegraphics[width=0.9\textwidth]{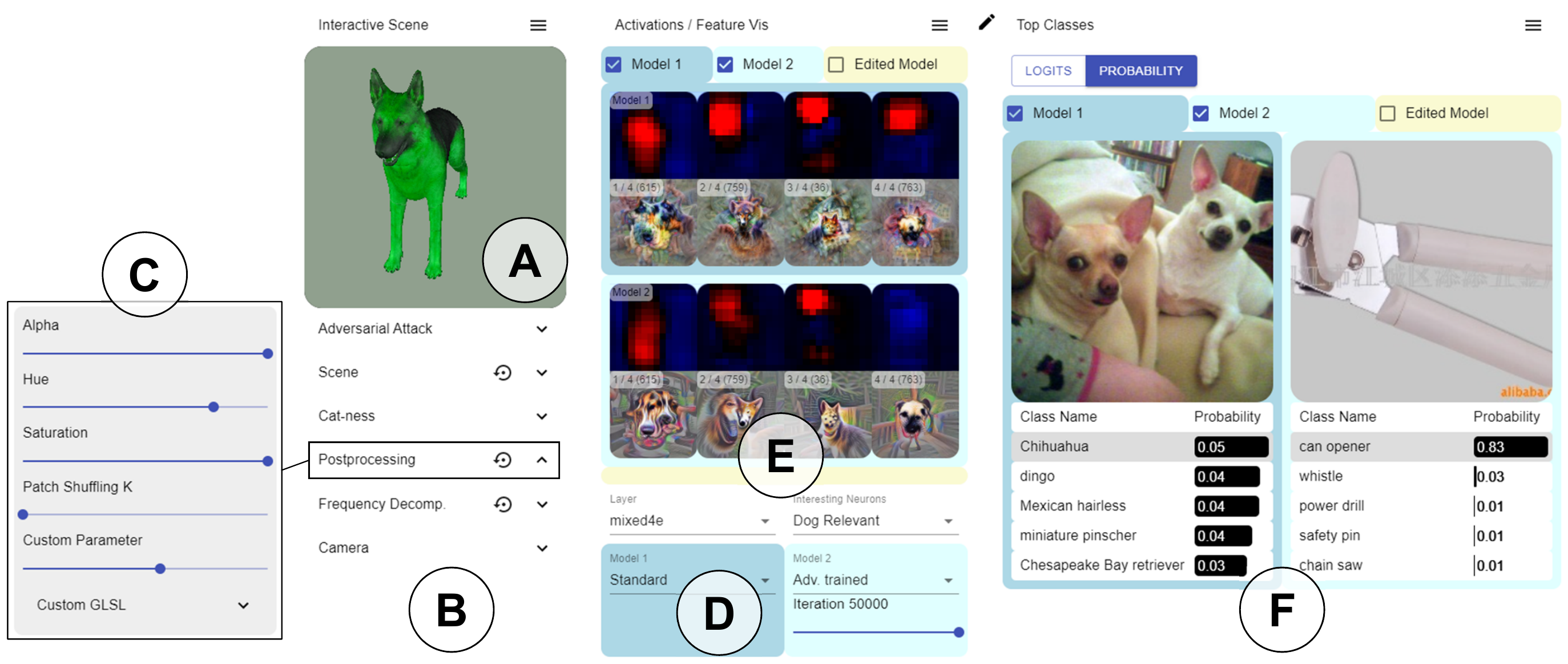}
\caption{Perturber interface: users manipulate the 3D scene (A) through a large variety of input perturbation methods that are grouped into functional categories (B). Per category, input perturbations can be seamlessly controlled through sliders (see inset (C)). Users can select two models to compare, as well as a set of neurons from dedicated layers (D). For the selected neurons and models, respectively, neuron responses are visualized live (E). Top-5 predictions for both models are shown as logits or probabilities along with an image example (F). }
\label{fig:interface}
\end{figure*}

Perturber supports these four core requirements through a highly interactive web-based playground consisting of the following components: the 3D input scene (Section~\ref{sec:inputScene}, Figure~\ref{fig:interface} A), the perturbation control (Section~\ref{sec:perturbationControl}, Figure~\ref{fig:interface} B-C), the prediction view (Section~\ref{sec:predictionView}, Figure~\ref{fig:interface} F), and the (comparative) neuron activation view (Section~\ref{sec:activationView}, Figure~\ref{fig:interface} E-D). These views allow for a qualitative inspection of the effects of input scene perturbations on one or two selected models (Section~\ref{sec:comparison}). 


Conceptually, Perturber can handle any CNN architecture. 
The current version supports models based on the Inception-V1 (also called GoogLeNet)~\cite{szegedy2015going} architecture. Our standard model was trained on ImageNet~\cite{russakovsky2015imagenet}. We use a pre-trained model with weights accessible through the \emph{Lucid}\footnote{https://github.com/tensorflow/lucid} feature visualization library~\cite{olah_feature_2017}. 

\subsection{Input Scene}
\label{sec:inputScene}

The input image is generated from an interactive 3D scene that can be manipulated in numerous ways (Section~\ref{sec:perturbationControl}). In computer vision projects, rendered images are often used instead of -- or in addition to -- photographs for training CNNs~\cite{su2015render}. 
Quantitative experiments have shown that, for high-level computer vision algorithms, the gap between synthetic and real images is fairly small~\cite{gaidon2016virtual}. CNN responses were shown to be consistent between simple rendered 3D models and natural images~\cite{aubry2015understanding}. Interactive exploration of CNN responses based on synthetic scenes therefore seems to be sufficiently representative of real-world applications. 
The major advantage of a synthetic 3D scene is that perturbation factors can be easily disentangled. That means, input modifications can be flexibly applied independently from each other to assess their isolated effects as well as their interactions. 


Out of the 1000 ImageNet classes, 90 of them are dog breeds. Therefore, there are numerous neurons in our Inception-V1 model that specialize in dog-related patterns, such as snouts, head-orientations, fur, eyes, ears, etc. As a consequence, we chose a dog as our main foreground 3D object to represent this special significance and to feed an input image that many hand-identified neuron circuits~\cite{cammarata_thread_2020} respond to strongly. As a second 3D model, we provide a vehicle, which has very different characteristics from a dog and is a commonly analyzed object in the machine learning community (e.g., \cite{gaidon2016virtual,aubry2015understanding}). \rev{In addition, users can upload custom 3D models.}
This way, users can verify observations they have made \rev{also with other models.}

\subsection{Perturbation Control}
\label{sec:perturbationControl}

The core principle of Perturber is that the user can flexibly vary the perturbation factors they want to explore (R1) and observe their effects instantly (R3). 
Below the input scene view, Perturber provides sliders for seamless control of the perturbations (Figure~\ref{fig:interface} C). Perturber provides more than 20 scene perturbation factors, as well as all possible combinations between these factors. 
We group these perturbation methods into the following categories: 

\textbf{Geometry} perturbations, such as rotation, translation, or cropping an image, are classic perturbations a CNN might be sensitive to~\cite{zheng2016improving,athalye_obfuscated_2018,engstrom2019exploring}. Therefore, affine image transformations are also a common data augmentation method for more robust training~\cite{perez2017effectiveness}. We support geometric perturbations through simple camera controls, which allow the user to arbitrarily orbit around the 3D model, as well as to freely zoom and pan the scene to create image cropping effects. In addition, users can rotate the camera around the z-axis to simulate image rotation. 

\begin{figure}[h]
\centering
\includegraphics[width=1\columnwidth]{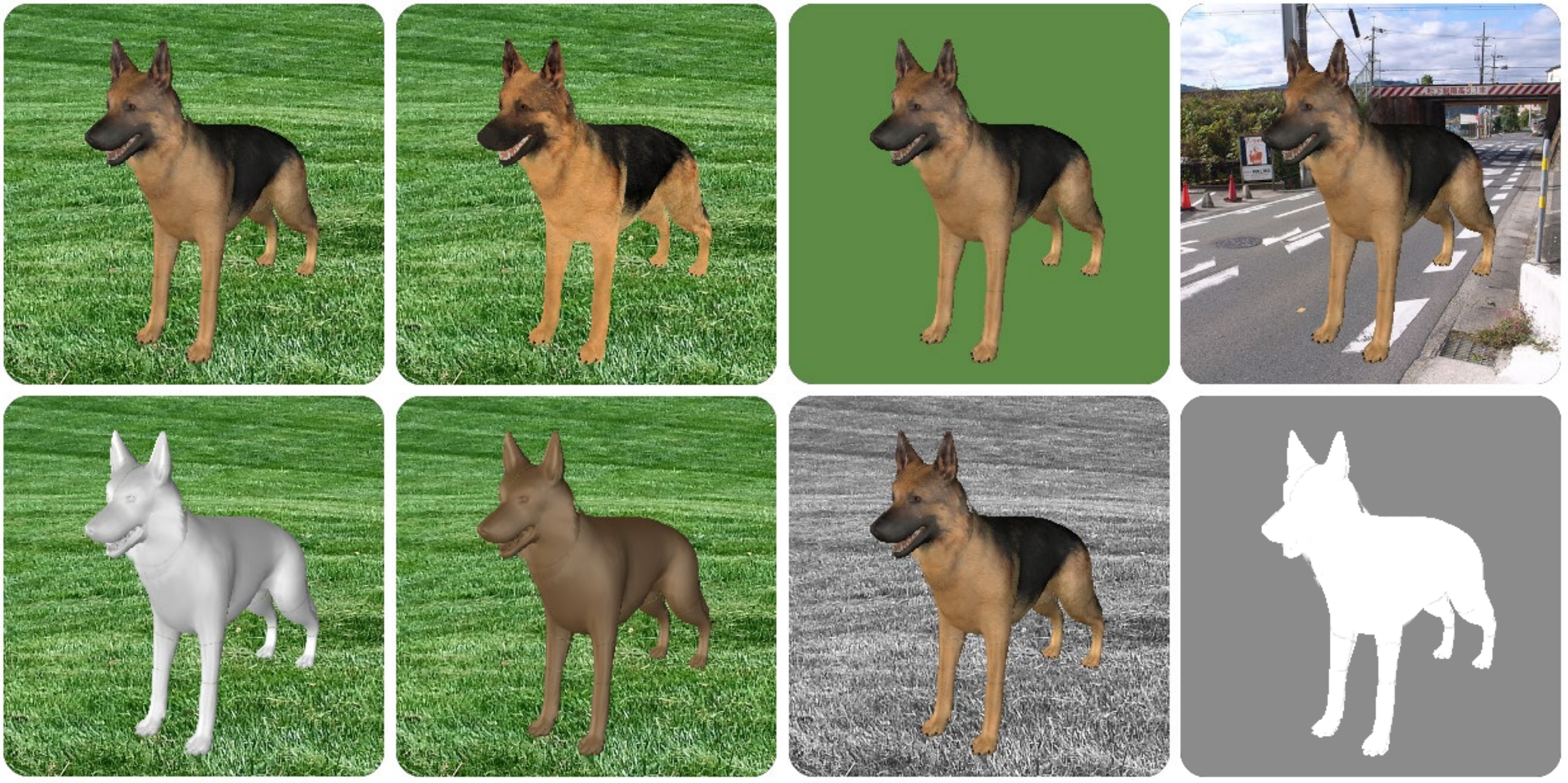}
\caption{Scene perturbations -- first row: original scene, no lighting influence, full background blur, different background image; second row: no texture influence, full texture blur, lowest background saturation, combination of background saturation, lighting influence, texture influence, and background blur. All perturbation parameters can be controlled seamlessly.  }
\label{fig:scenePerturbations}
\end{figure}

\textbf{Scene} perturbations that may seem irrelevant for human observers may easily confuse a standard CNN model. For example, changing the background~\cite{xiao2020noise} can have a tremendous effect on the prediction. We, therefore, support various scene perturbation operations, such as changing the background image, as well as modifying the lighting and texture parameters of the main object (see Figure~\ref{fig:scenePerturbations}). Through combinations of these parameters, specialized perturbations, such as silhouette images, can be generated (see Figure~\ref{fig:scenePerturbations} bottom right). 

\textbf{Shape} perturbations let users morph the shape of the main scene object into another one. Perturber currently supports morphing between dog and cat, as well as between a firetruck and a race car. By morphing the shape of an object independently from its texture, the texture-shape cue conflict~\cite{geirhos_imagenet_2019} can be investigated.

\textbf{Color} perturbations act on the rasterized 2d image. We support individual post-processing effects, such as alpha blending to black, hue shifting, and (de-)saturation. 
In addition to these parameters, users are provided with a text field where they can write their own GLSL code, taking a single parameter which is controlled by the slider. The code snippet defaults to code for contrast adjustment. Using these post-processing effects, users can assess the models' surprisingly high robustness against low contrast~\cite{dodge2016understanding} and low image saturation~\cite{sturmfels2020visualizing}.

\textbf{Frequency} perturbations selectively modify different image frequencies. Perturber provides three-parameter frequency decomposition that splits the image into low and high-frequency bands, which is achieved by Laplacian decomposition. Through selective frequency suppression, users can investigate phenomena such as the one described by Yin et al.~\cite{yin_fourier_2019}, where the authors show that adversarially trained networks are robust against high-frequency perturbations but very sensitive to low-frequency perturbations.


\textbf{Spatial} perturbations are post-processing effects systematically changing the image's pixel order. Perturber supports patch shuffling, which was used by Zhang and Zhu~\cite{zhang_interpreting_2019} to reveal a model’s sensitivity to global structure, which gets highly disturbed for human observers by patch shuffling. The image is divided into a grid of $k \times k$ cells, which are then randomly re-ordered. 

\textbf{Adversarial} perturbations, finally, are a core feature to understanding model robustness. Users can perform projected gradient descent (PGD) adversarial attacks~\cite{madry_towards_2019} on the current scene image. To do that, they choose a model to generate the attack from, a target class or the option to suppress the original prediction, as well as the attack $\epsilon$ and $L_p$ norm ($L_2$ or $L_{\infty}$). With each button press, the user performs one PGD step with a step length of $\epsilon/8$. This perturbation is costly and cannot be performed instantaneously. The first step typically takes around 10 seconds on a powerful consumer PC as the gradient function needs to be computed for the current input image first. To let users inspect the effect of an adversarial attack on the model fluidly like the other perturbation parameters, we overlay the current input image with the perturbation vector once it has been computed. This way, users can interactively fade the attack alpha using a slider and observe the system's response instantaneously. They can also fade the original image to inspect the perturbation vector itself. 

\subsection{Prediction View}
\label{sec:predictionView}

The prediction view shows the top-5 classification results for each model (Figure~\ref{fig:interface} F) either as probability or as logits. This allows the user to observe the classification changes resulting from input perturbations in real-time (R3). 
Each model's top result is represented by a class image example. Perturber shows the first image of the respective class in the ImageNet validation dataset. 


\subsection{Neuron Activation View}
\label{sec:activationView}

\begin{figure}[b]
\centering
\includegraphics[width=1\columnwidth]{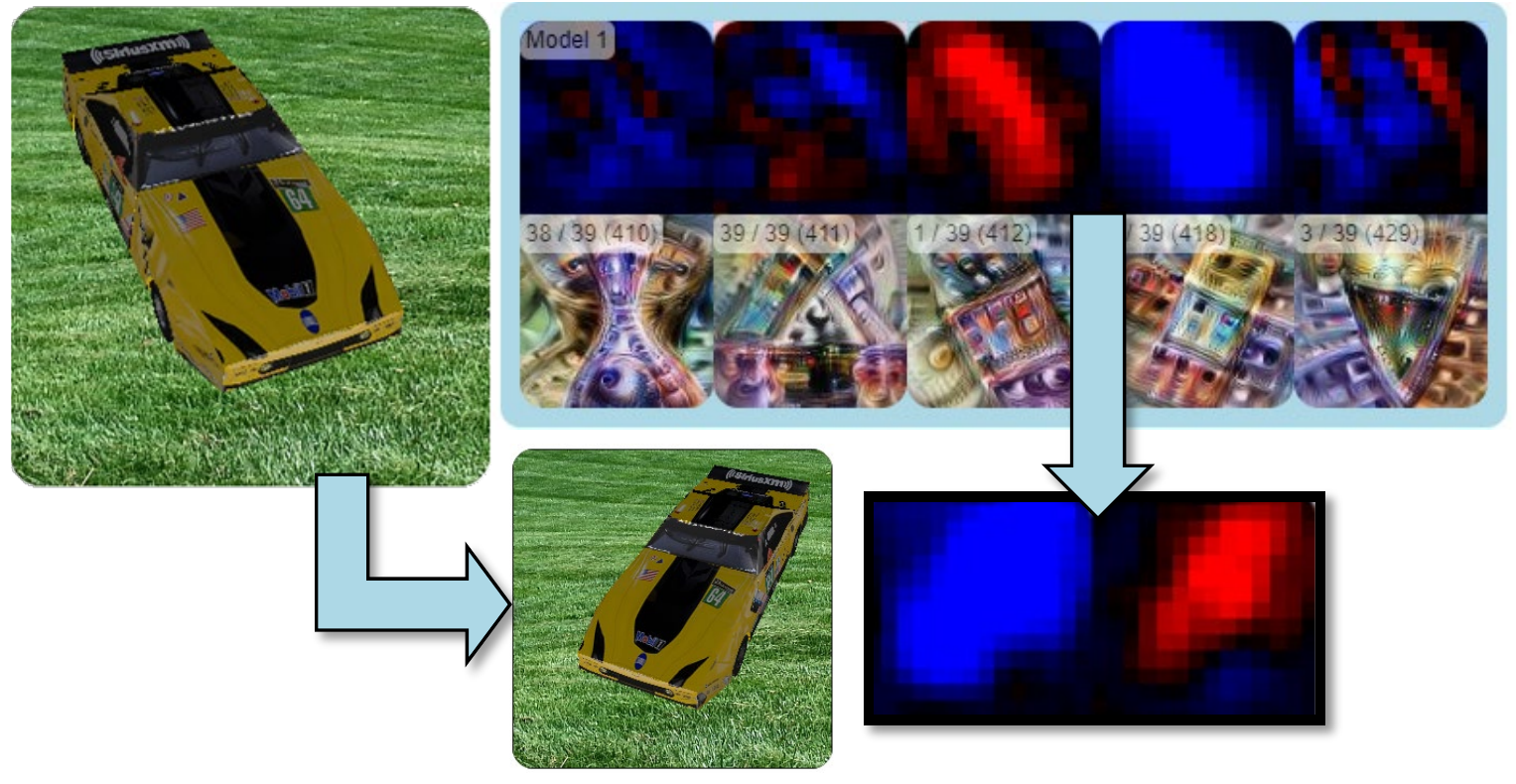}
\caption{Feature visualizations for neurons associated with complex shapes and curvatures in layer mixed4a for the standard model, as well as their activation maps for the input on the left. Note how rotating the input model causes an activation change for oriented shape detectors (insets on the bottom). }
\label{fig:neurons}
\end{figure}

The neuron activation view is the central interface for the analysis of how input perturbations affect the models' hidden layers (R2). Perturber represents neurons through feature visualizations. 
Feature visualizations aim at providing a lens into networks to visualize what patterns certain network units respond to \cite{olah_feature_2017}. These patterns might be, for example, edges in a particular direction in earlier layers (cf., Figure~\ref{fig:neurons}), or specific objects, like dog heads (cf., Figure~\ref{fig:teaser}) or car windows, in later layers. 

Activation maps show which regions of the input image cause the respective neuron to be activated. After experimenting with an implementation of the gradient-based visualization method \emph{Grad-CAM}~\cite{ribeiro_why_2016}, we decided to focus on the computationally far less expensive activation maps showing the neuron activations for a forward propagation pass. This way, neuron activations can be observed instantly (R3). 
Input regions that highly activate the neuron are shown in red, while blue regions cause a negative activation (see Figure~\ref{fig:neurons}). As the user manipulates the input scene, the activation maps update instantaneously so that the user can observe in real-time to which image features a neuron responds in one of the models. For example, in Figure~\ref{fig:neurons}, neurons 412 and 418 of layer mixed4a respond to oriented patterns. Rotating the textured race car causes these two neurons to strongly change their activations. 

Inception-V1 contains thousands of neurons. Clearly, it is impossible to show feature visualizations and interactive activation maps for all neurons concurrently. Instead, we provide pre-selected neuron groups and categories for each layer, which were characterized into semantically meaningful neuron families in the course of the \emph{Circuits} project~\cite{cammarata_thread_2020}. In addition to these previously presented neuron groups, we identified further sets of dog-, cat-, race car-, and firetruck-related neurons in later layers using \emph{Summit}~\cite{hohman_summit_2020}. These pre-selected neuron groups consisting of one to up to around 70 neurons can be selected from drop-down menus (Figure~\ref{fig:interface} D). 

\subsection{Model Comparison}
\label{sec:comparison}

After exploring potential sources of vulnerability, model designers commonly fine-tune existing models using augmented training data covering the identified vulnerability. To leverage Perturber for direct comparison of robustness (R4), we employ a transfer learning approach, which is initialized with the weights from the standard model. This is necessary to guarantee that feature correspondence of individual network units is kept intact. Only this way, we can assume that units across models respond similarly to identical input images. 

To illustrate what individual network units respond to, Perturber relies on feature visualizations. Feature visualizations are independent of the input image, but they differ between models. Indeed, Table~\ref{tab:featureVis} shows noticeable differences between a single neuron's feature visualizations of three different model variations. 

To showcase model comparison, Perturber already includes two robust model variations of Inception-V1: 
The first model was adversarially trained with projected gradient descent (PGD), which was described by Madry et al.~\cite{madry_towards_2019}. Adversarial training increases model robustness by incorporating strong adversarial attacks into the training procedure. The second model was trained by Stylized-ImageNet, which is a variation of ImageNet, where images were transformed into different painting styles using a style transfer algorithm~\cite{geirhos_imagenet_2019}. The purpose of Stylized-ImageNet was to induce a shape bias, while the standard ImageNet-trained models were found to be biased unproportionally towards texture~\cite{geirhos_imagenet_2019}. Users can choose two models for comparison from the interface (Figure~\ref{fig:interface} D). In addition, they can choose multiple checkpoints along the incremental fine-tuning process to analyze the development of the models during training. Corresponding neuron activation views are then juxtaposed in the Perturber interface for direct comparison (Figure~\ref{fig:interface} E). 

\section{Web-Based Implementation}
\label{sec:implementation}

Perturber runs purely on the client-side in the user’s browser and without any server-side computations at runtime. 
We precomputed all data that does not depend on interactively changeable elements to make the UI as efficient as possible. 
For transfer learning, we initialized the weights with those of the standard model (i.e., Inception-V1 trained on ImageNet), which were obtained through the \emph{Lucid} library~\cite{olah_feature_2017}. For adversarial fine-tuning, we used the open-source implementation by Tsipras et al.~\cite{tsipras_robustness_2019}. No layers were frozen for transfer learning. For both fine-tuned models, we used a learning rate of 0.003, a batch size of 128, and we trained until the models reached a top-5 training error of around 0.5, which took approximately 50K iterations for the adversarial fine-tuning and around 90K iterations for the Stylized-ImageNet fine-tuning.

To generate feature visualization, we employed the implementation provided by the \emph{Lucid}~\cite{olah_feature_2017} library.
For each model, we generated feature visualizations for 5808 neurons from the most relevant layers (i.e., the first three convolutional layers and the concatenation layers at the end of each mixed-block). 
During fine-tuning, we obtained feature visualizations for 17 checkpoints of adversarial fine-tuning and seven checkpoints of Stylized-ImageNet fine-tuning. 
For each neuron, we computed feature visualizations using two parametrization methods -- naive pixel or Fourier basis~\cite{olah_feature_2017} (see Table~\ref{tab:featureVis}). The second performs gradient ascent in Fourier space and leads to a more equal distribution of frequencies, resulting in more naturally looking feature visualizations. We use \emph{transformation robustness}~\cite{olah_feature_2017} in addition to both methods. Without Fourier basis parametrization, the differences between the models are more visually distinct (Table~\ref{tab:featureVis} top row).
The computation of all $\sim$300K generated feature visualizations required around one month on a machine with two NVIDIA GTX 1070 GPUs. \footnote{All generated feature visualizations can be downloaded from \url{https://github.com/stefsietz/perturber/}%
}. 

\begin{table}[]
\centering
\caption{Comparison between naive pixel (top) and Fourier basis  (bottom) parametrized feature visualizations of neuron 222 in layer mixed4a for three model variations.  }  
\label{tab:featureVis} 
\begin{tabular}{m{1.45cm}M{1.64cm} M{1.64cm} M{1.64cm}}
 & \textbf{standard} & \textbf{adversarial}  & \textbf{stylized} \\ 

 \textbf{naive pixel param.} & 
 \includegraphics[scale=.4]{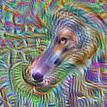} & \includegraphics[scale=.4]{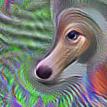} & 
 \includegraphics[scale=.4]{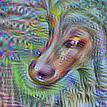} \\
 
 \textbf{Fourier basis param.} & 
 \includegraphics[scale=.4]{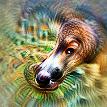} & \includegraphics[scale=.4]{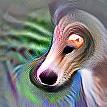} & \includegraphics[scale=.4]{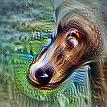}\\
\end{tabular}

\end{table}


Manipulation of the 3D scene, model inference based on the rendered image, and computation of activation maps are performed live in the web browser.  
The client relies on GPU acceleration for both, 3D scene rendering and CNN inference. We use the WebGL-based libraries \emph{Three.js} and \emph{TensorFlow.js}~\cite{smilkov_tensorflowjs_2019} for these tasks, respectively. The front-end GUI is based on \emph{React.js}. 
Input perturbations based on post-processing effects are implemented as multiple sequential render passes with custom GLSL shaders.
For model inference, we use \emph{Tensorflow.js}~\cite{smilkov_tensorflowjs_2019}, which enables fast GPU-accelerated CNN inference. \emph{Tensorflow.js} is also used for computing adversarial attacks. 

\begin{figure}[t]
\centering
\includegraphics[width=0.8\columnwidth]{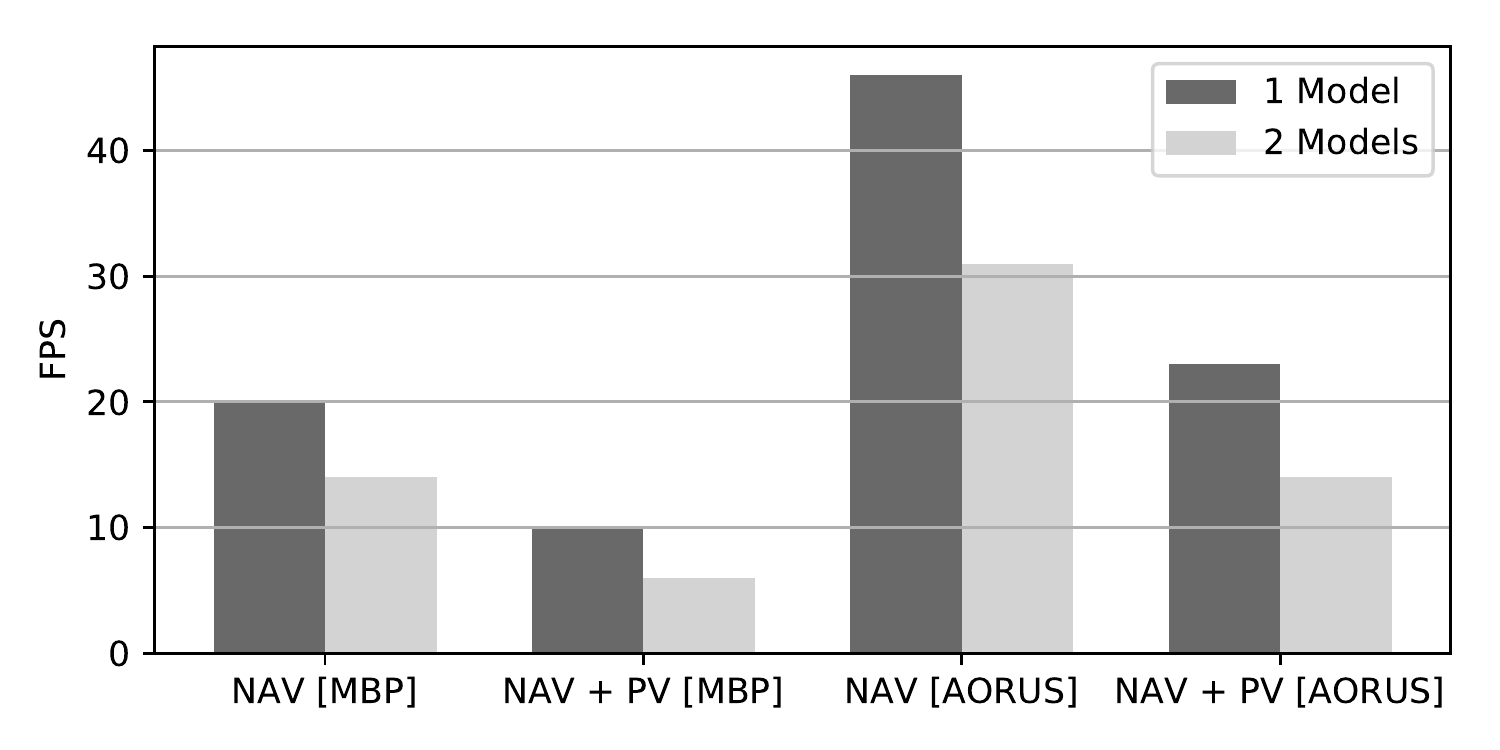}
\caption{Performance benchmarks for four different output configurations, measured on two machines (MBP or AORUS): neuron activation view (NAV) only, or neuron activation view in combination with prediction view (NAV+PV) visualized for one or two visualized models concurrently. }
\label{fig:performance}
\end{figure}

A major requirement of Perturber is that the effects of input perturbations can be observed instantaneously (R3). 
To assess requirement R3, we measured the client's performance while constantly orbiting the camera around the object. Figure~\ref{fig:performance} shows the recorded frame rates for two client notebooks: a MacBook Pro 13” 2018 with Intel Iris Plus Graphics 655 (MBP) and an AORUS 15G Gaming Notebook with an NVIDIA GeForce GTX 2080 Super GPU (AORUS). It is clearly visible that the GPU of the client machine has a strong influence on the frame rate. The application performance also depends on the enabled visualizations, with the prediction view (Section~\ref{sec:predictionView}) being more computationally expensive than showing four live activation maps (Section~\ref{sec:activationView}). When using less powerful client machines, users can selectively switch off views.

\section{Exploration Scenario: Texture-Shape Conflict}
\label{sec:scenarios}

To demonstrate the capabilities of Perturber, we selected one important aspect of robustness: the influence of shape and texture on CNN classification results. It has been shown that CNNs are biased towards texture, which affects robustness~\cite{geirhos_imagenet_2019}. In this scenario, we first aim to investigate whether this effect could have been discovered using Perturber. Secondly, we aim to qualitatively validate whether a more robust model variation, which was fine-tuned using Stylized-ImageNet~\cite{geirhos_imagenet_2019}, can improve upon this bias.  

\begin{figure}[h]
\centering
\includegraphics[width=0.7\columnwidth]{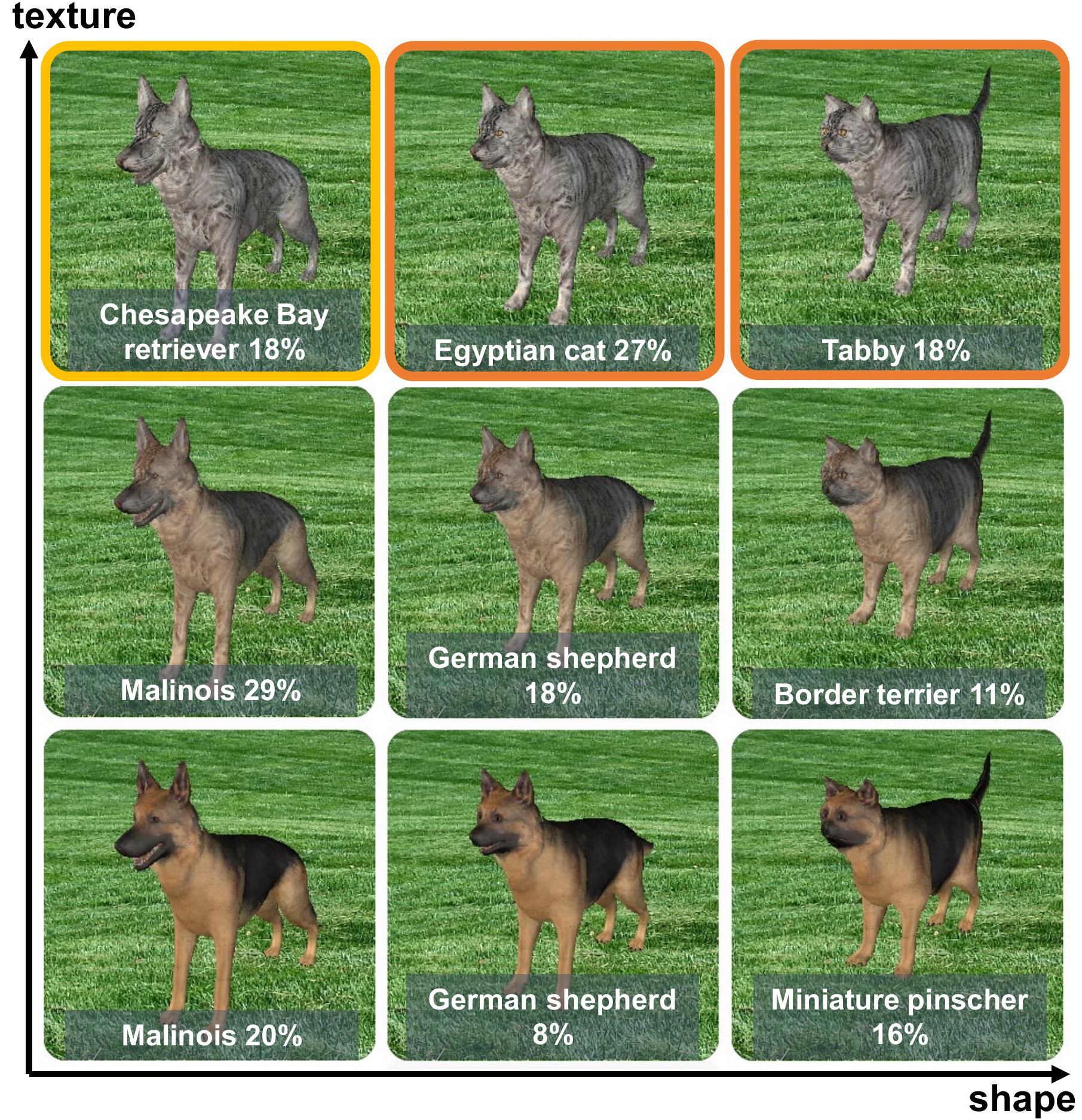}
\caption{Object morphing between dog (bottom left) and cat (top right) along the texture dimension (y axis) and shape dimension (x axis): Top-1 standard model predictions with their probabilities are shown as text labels. Input images leading to cat breeds within the top-5 or top-1 predicted classes are indicated by a yellow and orange frame, respectively (top row).  }
\label{fig:morphing}
\end{figure}

For our first analysis, we explore the texture-shape cue conflict through shape and texture perturbations. Figure~\ref{fig:morphing} illustrates that shape perturbations alone do not cause the standard model to predict a cat breed. However, when morphing the texture, the model predictions switch to cat breeds -- even when the object shape remains unchanged. This is a first indicator that the model is indeed much more sensitive to texture than to shape perturbations. 

\begin{figure}[]
\centering
\includegraphics[width=0.8\columnwidth]{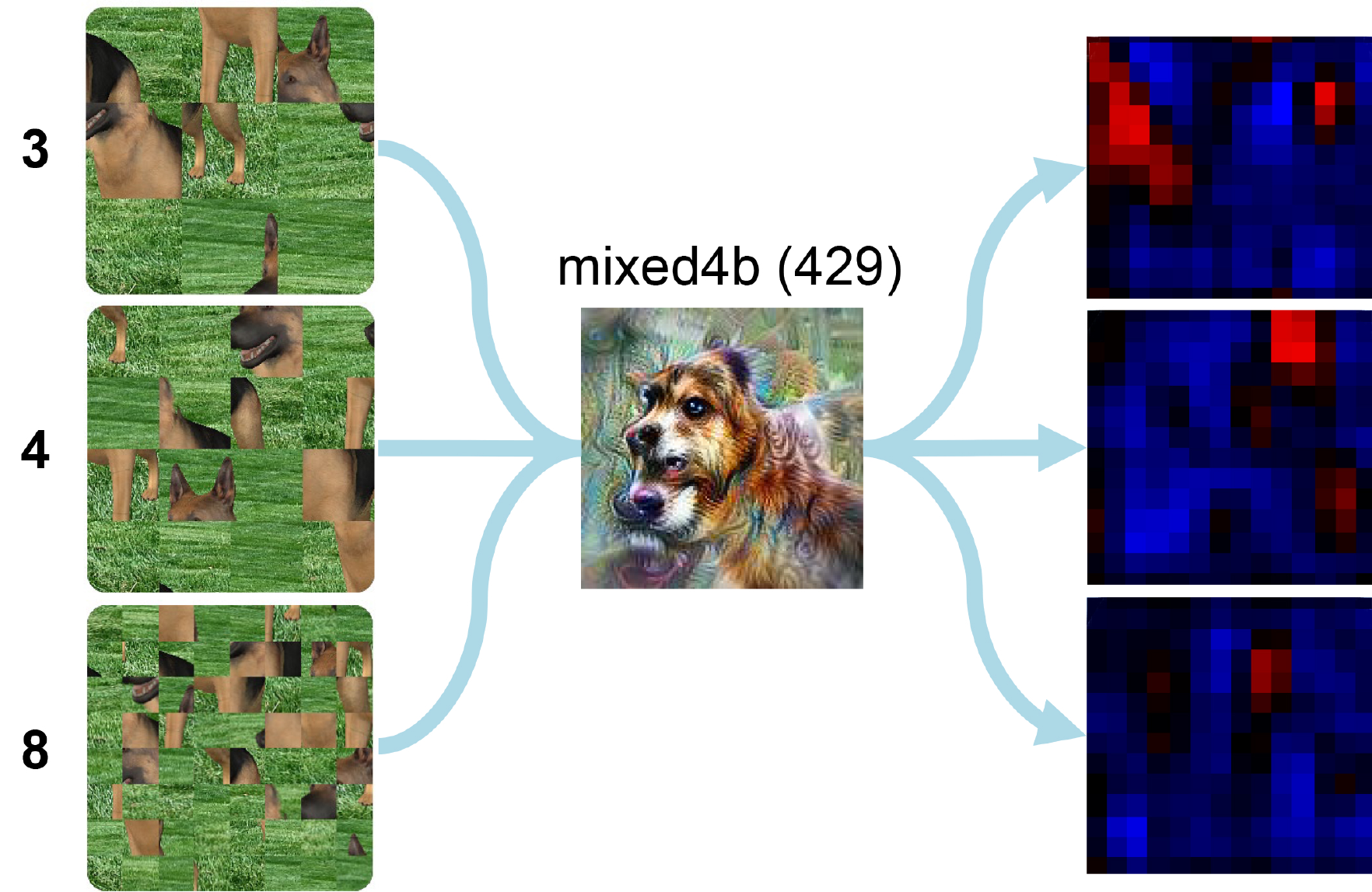}
\caption{Randomly shuffling the dog scene into 3 (top), 4 (middle), and 8 (bottom) patches: A dog-related neuron in mixed4b of the standard model is activated by patches containing parts of the dog's face and textured body parts.  }
\label{fig:patchShuffling}
\end{figure}

To further investigate the texture sensitivity of the standard model, we replicated the patch shuffling experiment by Zhang and Zhu~\cite{zhang_interpreting_2019}. Using Perturber, we can inspect single neurons' activations during this experiment. We illustrate our observations on a hand-picked neuron in Figure~\ref{fig:patchShuffling}, which is strongly activated by dog faces looking to the left. Note how this neuron is activated by strong texture contrasts, especially around the mouth of the dog. Image regions containing ears do not activate this particular neuron. For this scene, the standard model still predicts a dog breed up to $k=7$ randomly shuffled image patches. This illustrates that the decomposed shape has indeed very little influence on the model prediction.  



\begin{figure}[h]
\centering
\includegraphics[width=0.7\columnwidth]{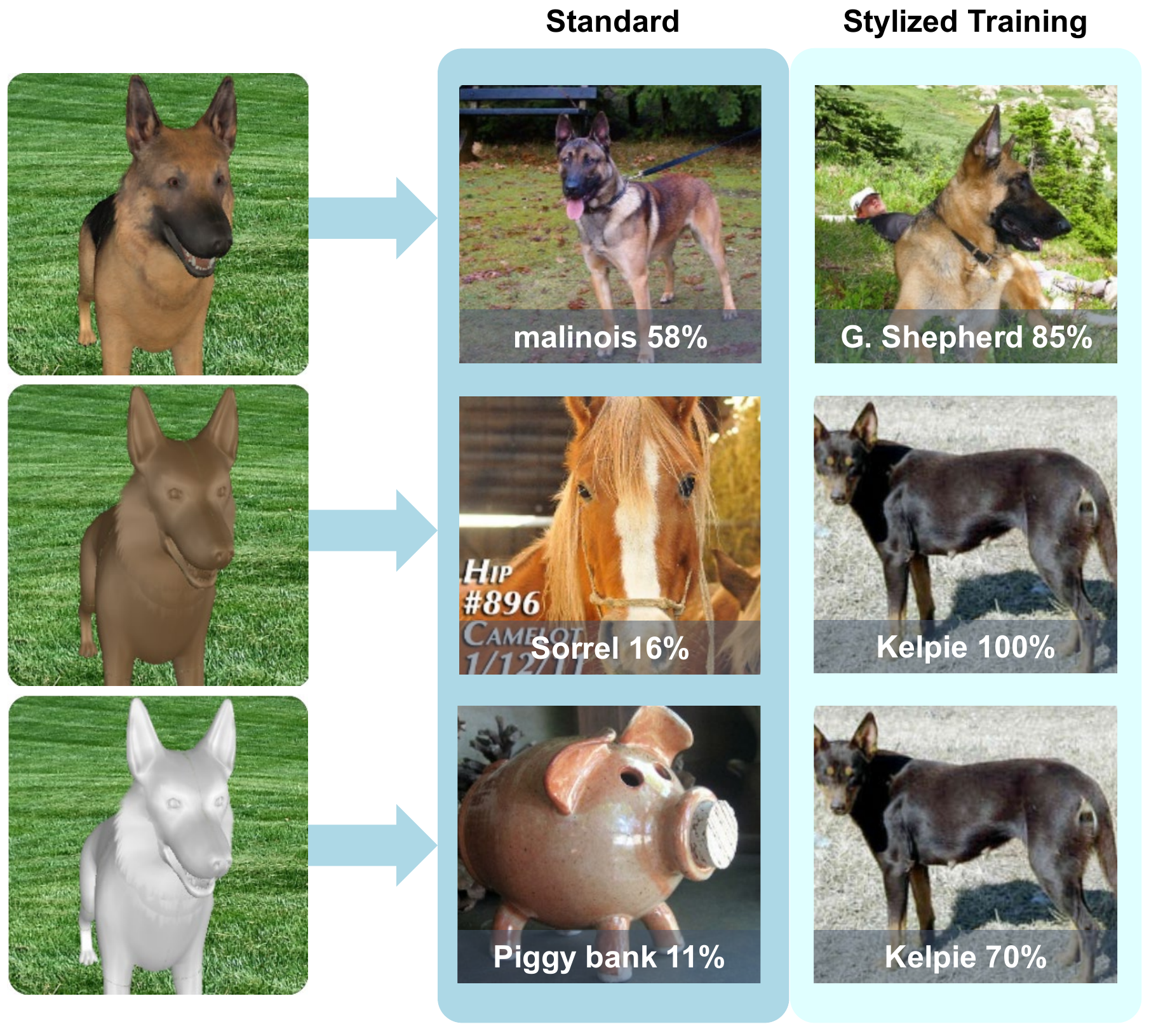}
\caption{Inspecting the shape vs.~texture conflict through scene perturbations for the standard (left) and the Stylized-ImageNet trained model (right): while the standard model gets confused by missing textures, the Stylized-ImageNet trained model predicts a dog breed even if the 3D model is untextured.}
\label{fig:stylizedTexture}
\end{figure}

In the next step, we analyze if the standard model is indeed more sensitive to texture than the model fine-tuned on Stylized-ImageNet~\cite{geirhos_imagenet_2019} by gradually removing the texture of the main object. The predictions in Figure~\ref{fig:stylizedTexture} confirm that the Stylized-ImageNet trained model consistently predicts a dog breed, even in the absence of a texture. The standard model, on the other hand, seems to rely much more on the texture. The model trained with Stylized-ImageNet is also more sensitive to patch shuffling, which is another indication that it relies less on texture information than the standard model (see Section A in the Supplemental Document for image examples).  

\begin{figure}[b]
\centering
\includegraphics[width=1\columnwidth]{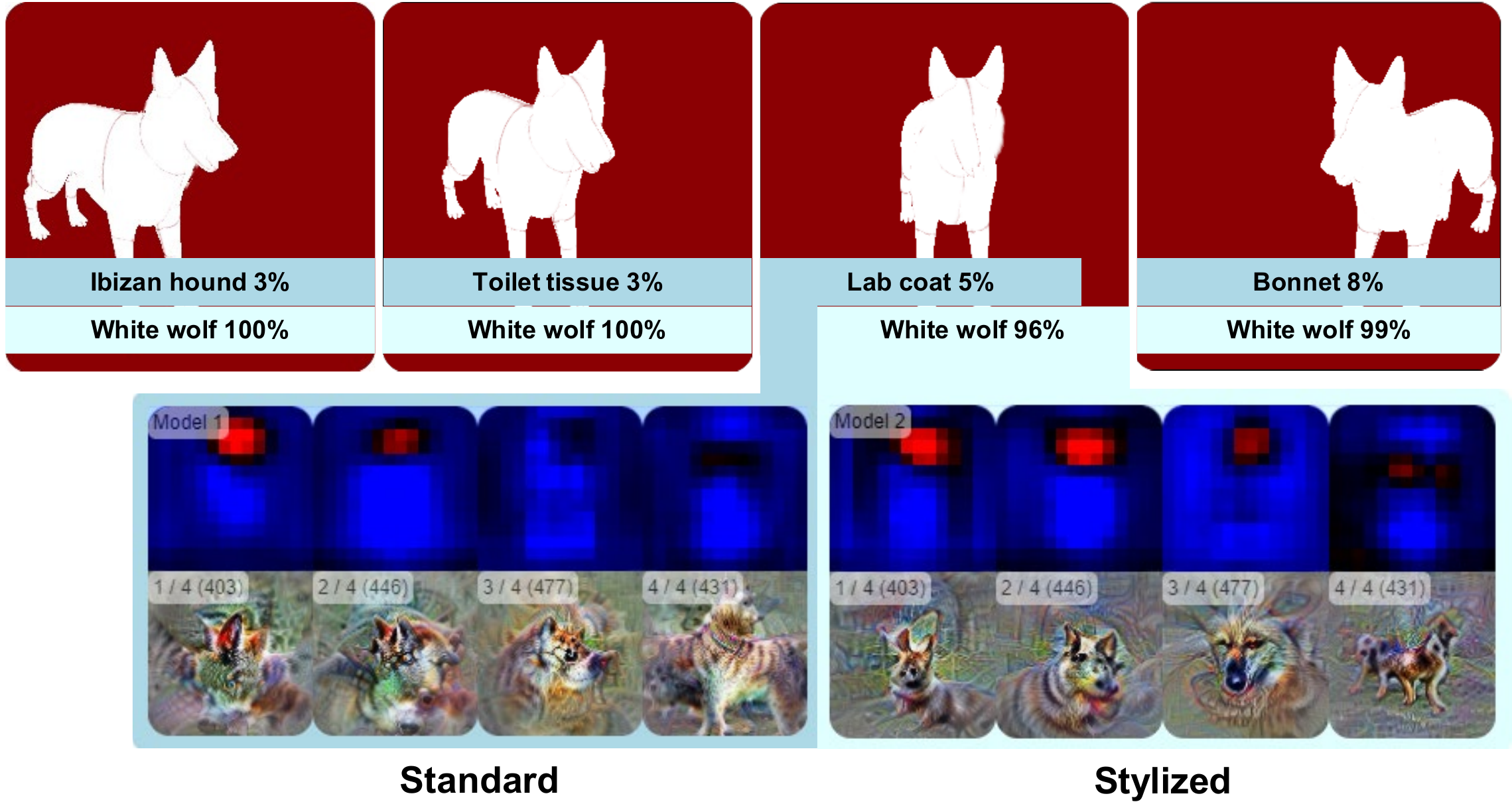}
\caption{Rotating a dog silhouette: the corresponding predictions of the standard model and the model trained on Stylized-ImageNet, as well as activations of dog-related neurons for the front-facing dog image in mixed4d of the standard model (left) and the robust model (right). }
\label{fig:silhouette}
\end{figure}

Finally, we compare shape sensitivity between the two models. To this end, we combine various scene perturbations to generate a silhouette image of the dog. We then gradually change the pose of the dog. Figure~\ref{fig:silhouette} shows the predictions of the standard model and the model fine-tuned with Stylized-ImageNet. Clearly, the predictions of the standard model are unstable, especially when the dog is directly facing the camera. The model trained with Stylized-ImageNet, however, reliably predicts a white wolf with high probability. The activations of relevant neurons indicate that the Stylized-ImageNet-based model has learned a more stable representation of a frontal dog head, which also get activated by a silhouette image. 

These examples illustrate that Perturber provides flexible ways to explore and better understand potential threats to robustness, such as the texture-shape conflict of CNNs. Section A in the supplemental document contains more exploration scenarios.

\section{Case Studies}
\label{sec:caseStudies}

While the previous scenario investigated a known vulnerability, we further explored whether Perturber can help to discover unknown threats to robustness through case studies with machine learning experts. 
We conducted case studies with five machine learning researchers (four PhD students, one post-doctoral researcher; one female, four males). Except for one user, sessions were conducted individually through video conferencing and lasted approximately one hour each. One user preferred to perform the case study offline and sent a textual text report instead. 

The researchers cover various topics of expertise in the field of AI interpretability and the design of robust machine learning models. The concrete research areas are listed in Section B of the supplemental document. 
Two users were involved in the co-design process of Perturber and are co-authors of the paper. Three users were unfamiliar with the system before the evaluation. One of these users entered the co-design iteration process after the case study and is also a co-author. Paper co-authorship is a common collaboration role in design studies~\cite{sedlmair2012design}. 

During the video conference, the participant used the online tool while sharing his/her screen with the first and last author. Every session was recorded while conversations and observations were transcribed on-the-fly or in retrospective through automatic speech-to-text. 

Every video conferencing session started with a short introduction by the participant describing his/her research focus. Afterwards, we gave a short demonstration of Perturber's features. We then asked the participant to shortly comment on his/her first impression and his/her expected insights from the analysis. Then, the participant freely played around with Perturber while thinking aloud. In particular, we asked the user to always state his/her intent before performing an action and whether he/she would have any particular hypotheses about the response and behavior of the network based on the chosen input. If a user could not find the respective functionality of the tool, the first and/or last author provided oral assistance. At the end of the study, the user was encouraged to summarize his/her impressions, the potential benefits of the tool, and to provide suggestions for improvements.

\subsection{Observations and Feedback}

\begin{figure}[h]
\centering
\includegraphics[width=0.9\columnwidth]{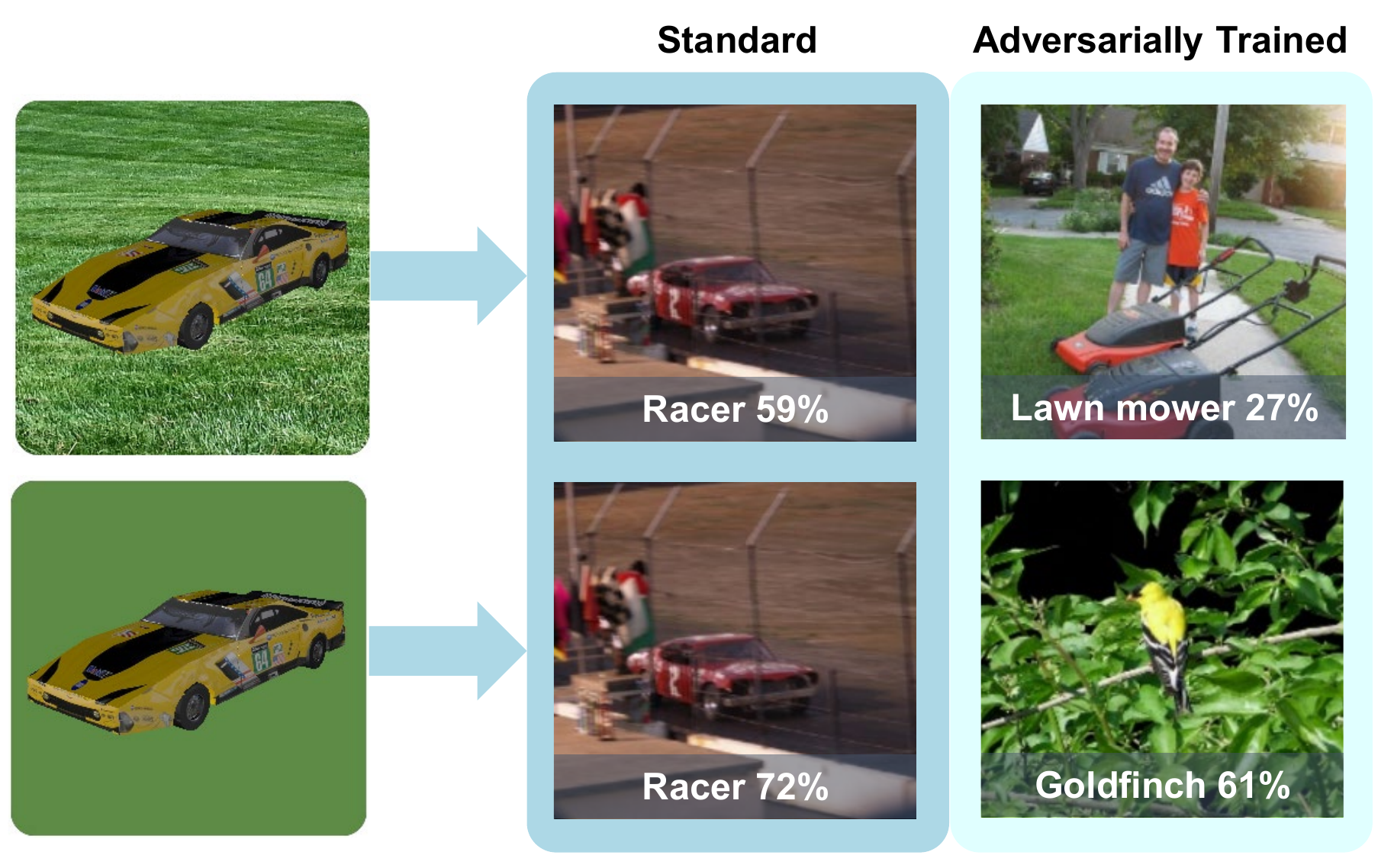}
\caption{Blurring the background of the race car has little influence on the standard model's predictions (left). The adversarially trained model (right) is very sensitive to the chosen background. The probability for ``racer'' is 11\% with a grass background and 7\% with a uniform green background. }
\label{fig:backgroundBlurPredictions}
\end{figure}

Users praised the fact that Perturber works ``live'' and therefore allows for ad-hoc \textbf{exploratory analyses}. They liked that Perturber allows to play around with simple examples to quickly find patterns and form hypotheses. Generally, the main focus of the exploratory analyses was trying to identify input perturbations where a model would respond unexpectedly. One user described this process as trying to answer the following questions: \emph{``How can I break a model? What do I need to do so that the resulting prediction is wrong?''} For example, three users were surprised to see how vulnerable the adversarially trained model seems to be to some geometric changes, such as zooming or rotating. Two users also discovered a high sensitivity of adversarially trained models to background modifications, as illustrated in Figure~\ref{fig:backgroundBlurPredictions}. 
The live approach was praised in particular for cases without a clear ground truth, such as object morphing (Figure~\ref{fig:morphing}). Such scenarios would not be possible to assess quantitatively without human subject studies. Being able to quickly generate hypotheses that could then be formally tested in a more controlled setting was considered very useful. 


\begin{figure}[b]
\centering
\includegraphics[width=1\columnwidth]{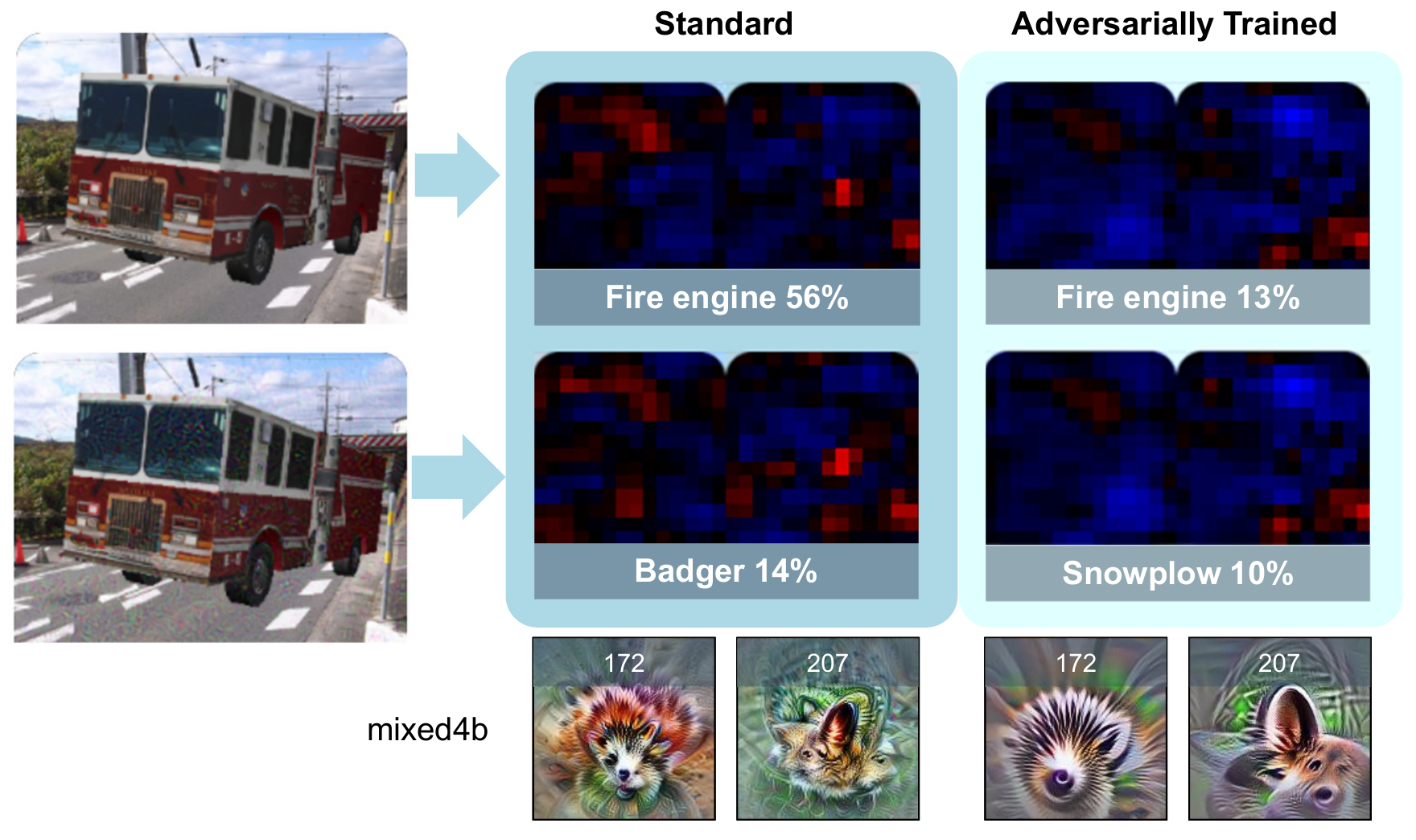}
\caption{Adversarial attack with target class ``badger'' on the fire truck scene (top) and the effect on two cat-related neurons in layer mixed4b for the standard model (left) and adversarially trained model (right): As the attack strength is increased (bottom), activations of some standard model neurons increase while the activations of the adversarially trained model are hardly affected. The prediction probabilities for the fire engine decrease to 0.1\% for the standard model and to 6\% for the adversarially trained model after the attack. }
\label{fig:adversarialAttack}
\end{figure}

We also observed indications that Perturber is practically helpful for \textbf{visual confirmation}. 
For example, using the model predictions and activations of selected neurons, one user verified that adversarial attacks only affect the standard model. He also observed how animal-related neurons of the standard model got activated during an attack with the target class ``badger'' (Figure~\ref{fig:adversarialAttack}). 
Not all assumptions were actually confirmed. For example, one user expected that the model trained on Stylized-ImageNet would be noticeably more vulnerable to image blur than the standard model. Unexpectedly, the model's responses were not more sensitive to blur than the standard model's for the chosen input scene (see Section A in the supplemental document). 

Finally, a user pointed out that playing around with Perturber would \textbf{let non-experts get an intuition} how easily CNNs are fooled. For example, one user demonstrated how the standard model sometimes rapidly changes its predictions during a simple translation of the dog. Another expert demonstrated that a rotation of the dog along the z-axis in combination with an unusual background (street) was sufficient to disturb the adversarially trained model. Accordingly, he stated that Perturber could be informative for ``people fearing that AI will take over the world''.

A comprehensive list of observations reported by the individual users can be found in Section B of the supplemental document. 
Overall, the most frequently performed perturbations leading to the most interesting observations in our case studies were 1) geometric transformations, such as object rotation, zooming, and translation, 2) modification of the background, 3) combinations of (1) and (2), as well as 4) object morphing. The prediction view was perceived as giving instant, easily interpretable, and useful feedback. It was thus the primary view to observe model behavior. Feature visualizations were considered useful to characterize the difference between the models. Participants described feature visualizations of the adversarially trained model as more ``intuitive'' or ``cartoonish'' compared to the corresponding standard model's neurons. One participant found feature visualizations sometimes hard to interpret. One recommendation therefore was to additionally show strongly activating natural image examples from a dataset. 

\subsection{Quantitative Evaluation of User Observations} 

We performed quantitative measurements to see whether what users observed visually can be generalized beyond the given input scene, synthesized input images, and the Inception-V1 network architecture. To test the generalizability of the users' observations, we performed quantitative measurements using different models than the ones used in the online tool. Specifically, we used the pre-trained ResNet50 from the \emph{torchvision} library of \emph{PyTorch} \cite{pytorch} as the standard model. For comparison, we used weights of an adversarially trained version of the same model from the \emph{robustness} library~\cite{robustness} (ResNet50 ImageNet $L_2$-norm $\epsilon$ 3/255).

First, we investigated the adversarially trained model's sensitivity to background changes, which was reported by two users in the case study. For verification, we performed the \emph{Background Challenge} by Xiao et al.~\cite{xiao2020noise}, where a model is tested with (natural image) adversarial backgrounds. 
 The adversarially trained model can only correctly predict 12.3\% under background variations. The standard model achieves 22.3\% accuracy. Using this test dataset, random guessing would yield 11.1\% accuracy~\cite{xiao2020noise}. This shows that the robustified network is considerably more susceptible to adversarial backgrounds than the standard model.

\begin{figure}[]
\centering
\subfloat[]{
\includegraphics[width=0.9\columnwidth]{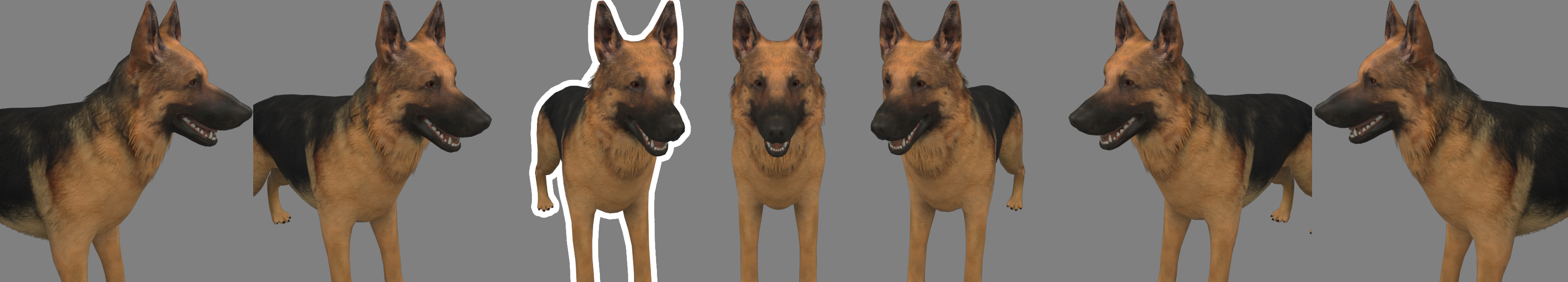}
\label{fig:yaw}
}
\vspace{-0.2cm}
\\
\subfloat[]{
\includegraphics[width=0.9\columnwidth]{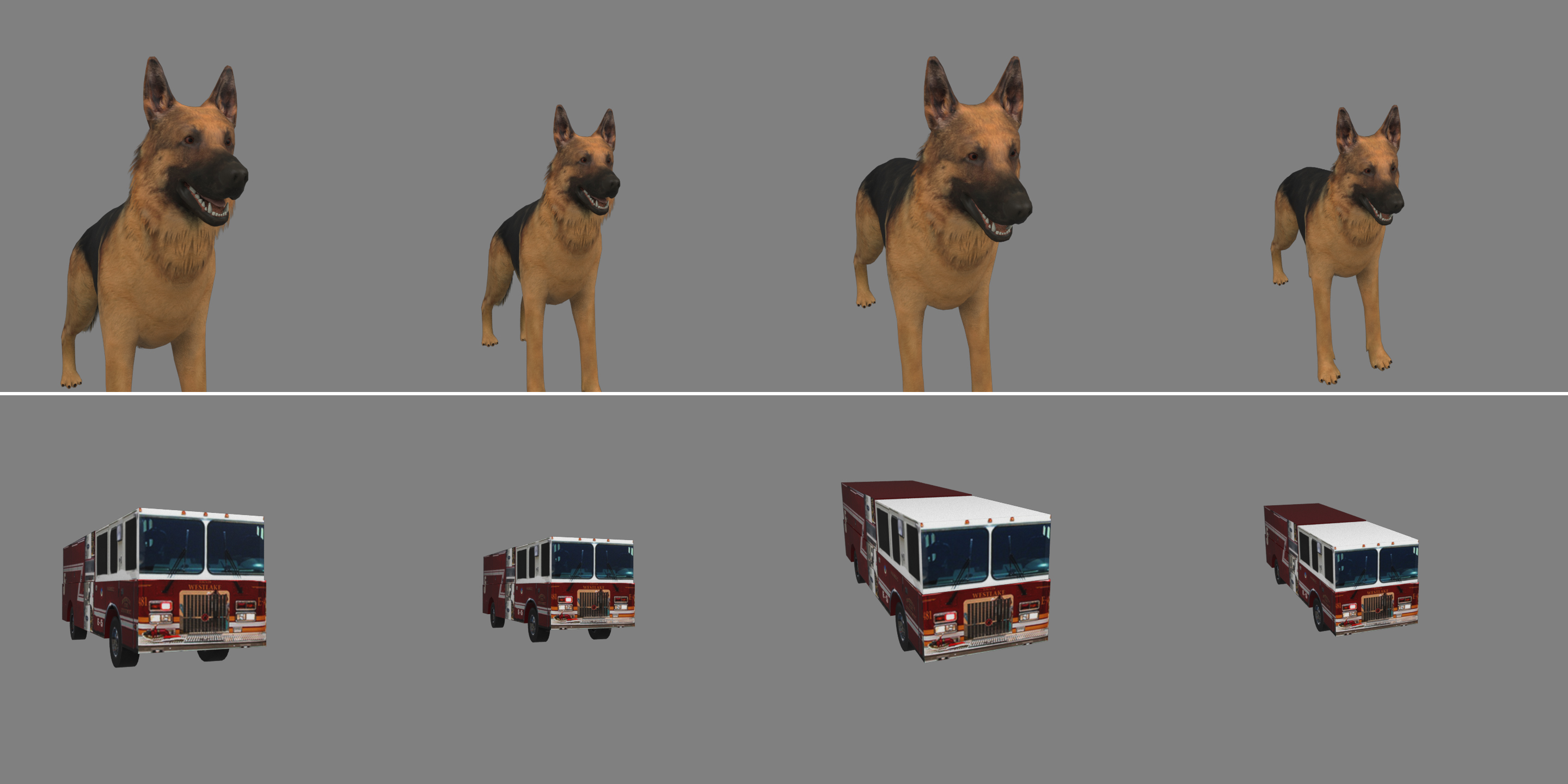}
\label{fig:prototypes}
}
\caption{Seven yaw angles variations (a) tested for four prototype views (b) per model (two of the four models are shown here). The prototype view in (a) is highlighted.} 
\label{fig:rotation}
\end{figure}

Second, we tested if the adversarially trained model is indeed more sensitive to geometric scene transformations than the standard model (reported by three users). In particular, we looked into camera rotation. We generated a synthetic dataset, where we rendered the four 3D models provided in Perturber from seven yaw angles, ranging from -70\degree to 70\degree\ (Figure~\ref{fig:yaw}), two pitch angles, and two distances of the camera to the object, as shown in Figure \ref{fig:prototypes}. In total, we generated 28 views for each of the four 3D models. 

To compare how much the predictions fluctuate, we chose a prototype view with a yaw angle of -23.3\degree for each of the four pitch / distance combinations and 3D model (Figure \ref{fig:prototypes}). For each of the 16 resulting prototypes, the logits of the top-10 classes served as ground truth vector $\vec{l}_{10}^{\star}$. 
For the 16 prototype views, we then calculated a fluctuation score $f_p$:

\begin{equation}
f_p = \sum_{y} \frac{\lVert \vec{l}_{10}^{\star} - \vec{l}_{10}^{y} \rVert_2}{s(\vec{l}_{n}^{\star})}, 
\end{equation}

where $\vec{l}_{10}^{\star}$ are the top-10 predictions of the prototype view, $\vec{l}_{10}^{y}$ is the logit vector of these 10 classes for the view associated with yaw $y$, and $s(\vec{l}_{n}^{\star})$ is the standard deviation of the logits of all $n$ classes in the prototype view. In other words, the fluctuation score measures how strongly the logits of the top-10 prototype classes diverge in the rotated input images. 

\begin{figure}[h]
\centering
\includegraphics[width=0.8\columnwidth]{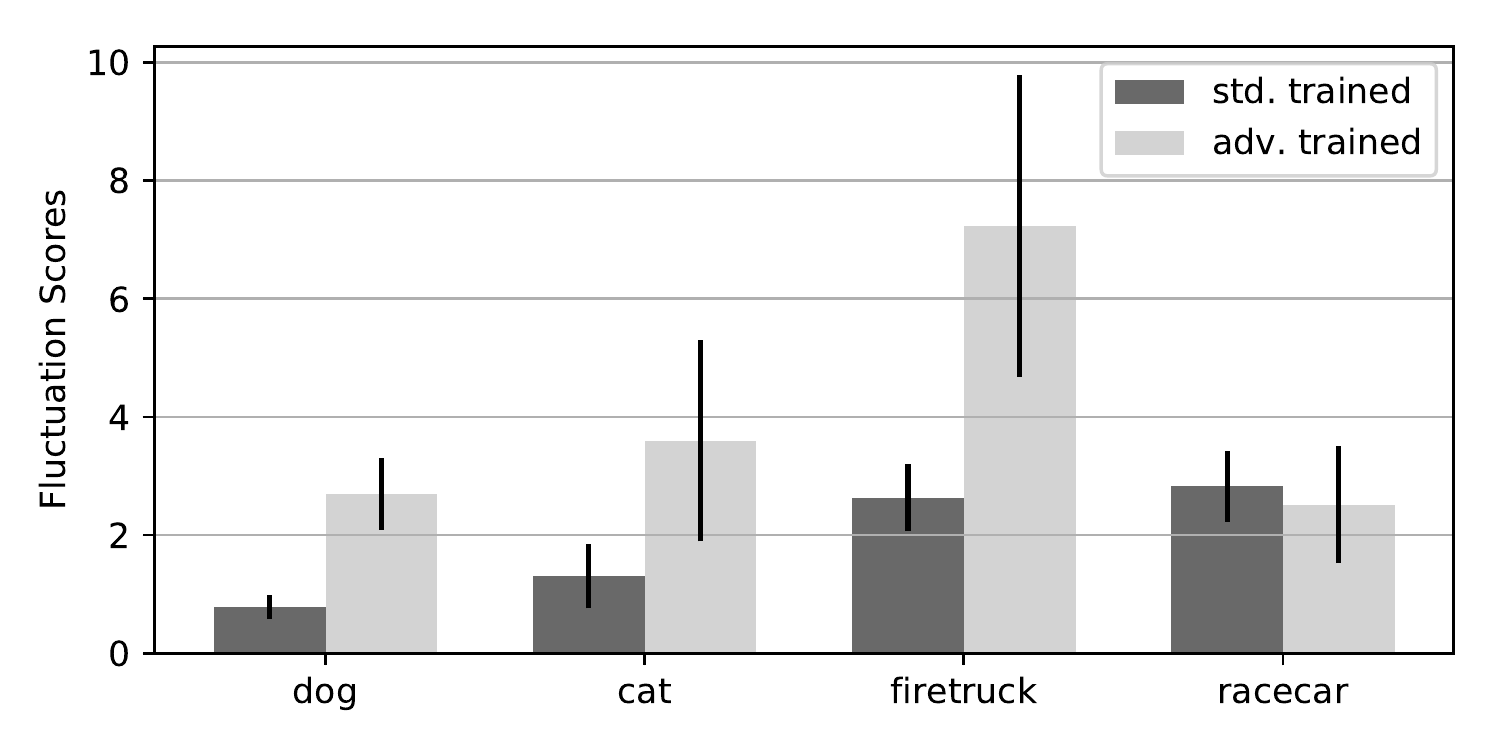}

\caption{Average yaw fluctuation scores for the standard model and the adversarially trained model across the four 3D models. Error bars show the standard deviation. }
\label{fig:score}
\end{figure}

Figure~\ref{fig:score} shows the average fluctuation score of all 28 prototype views. 
The fluctuation scores are considerably higher for the adversarially trained model for three of the four 3D models. This verifies that the adversarially trained model can be indeed more vulnerable to rotations of the main object. 

\section{Discussion \& Conclusions}


We showed that interactive perturbations in combination with live activations can be an effective method to explore potential vulnerabilities of CNNs and to perform qualitative evaluations of more robust model variations. In an exploration scenario, we could replicate the known texture-shape conflict~\cite{geirhos_imagenet_2019} through multiple perturbation examples. Machine learning experts participating in our case study observed a variety of known but also unexpected network behaviors. \rev{They could successfully replicate known CNN properties, such as models’ varying sensitivity to adversarial attacks or patch shuffling, using our synthetic input scenes. In addition, our quantitative post-study experiments of selected user observations could replicate their observations using a different network architecture and natural images. }
Through these experiments, we demonstrated vulnerabilities of adversarially trained models to background modifications and yaw rotations of the main object that, to the best of our knowledge, have not been discussed yet in the machine learning community. 

The majority of insights reported by our users were based on observations how the model predictions changed when perturbing the image. This implies that the principle of live input perturbations could also be useful for pure black-box models. To get a better intuition about which image features are influential for the model's final decision, our exploration scenarios indicate that neuron activations are also essential. But due to the large number of logical neuron units distributed among multiple layers, it can be time-consuming to find a set of neurons that eventually is affected strongly by the performed input perturbation. In the future, we thus plan to investigate alternative methods to select the displayed neurons in the neuron activation view. Also visual guidance to support the discovery of potentially harmful perturbation factors would be helpful. However, traditional guidance mechanisms may require costly computation, which will hamper interactivity (R3). Effective guidance mechanisms can therefore be considered interesting future work. \rev{Another interesting line of future work would be the support for encoder-decoder architectures, used prominently for semantic segmentation and image translation tasks among others. This could be facilitated by replacing the prediction view with a continuously updating display of the generated image.}  

\section*{Acknowledgments}

We thank Robert Geirhos and Roland Zimmermann for their participation in the case study and valuable feedback, Chris Olah and Nick Cammarata for valuable discussions in the early phase of the project, as well as the Distill Slack workspace as a platform for discussions.
M.L.~is supported in part by the Austrian Science Fund (FWF) under grant Z211-N23 (Wittgenstein Award).
J.B.~is supported by the German Federal Ministry of Education and Research (BMBF) through the Competence Center for Machine Learning (TUE.AI, FKZ 01IS18039A) and the International Max Planck Research School for Intelligent Systems (IMPRS-IS). R.H. is partially supported by Boeing and Horizon-2020 ECSEL (grant 783163, iDev40).


\bibliographystyle{eg-alpha-doi}  
\bibliography{main}        


\end{document}